# GeoAI at ACM SIGSPATIAL: The New Frontier of Geospatial Artificial Intelligence Research

Dalton Lunga[1], Yingjie Hu[2], Shawn Newsam[3], Song Gao[4], Bruno Martins[5], Lexie Yang[1], Xueqing Deng[3]
[1]Oak Ridge National Laboratory, USA
[2]University at Buffalo, USA
[3]University of California, Merced, USA
[4]University of Wisconsin-Madison, USA
[5]University of Lisbon, Portugal

**Abstract**

*Geospatial Artificial Intelligence (GeoAI) is an interdisciplinary field enjoying tremendous adoption. However, the efficient design and implementation of GeoAI systems face many open challenges. This is mainly due to the lack of non-standardized approaches to artificial intelligence tool development, inadequate platforms, and a lack of multidisciplinary engagements, which all motivate domain experts to seek a shared stage with scientists and engineers to solve problems of significant impact on society. Since its inception in 2017, the GeoAI series of workshops has been co-located with the Association for Computing Machinery International Conference on Advances in Geographic Information Systems. The workshop series has fostered a nexus for geoscientists, computer scientists, engineers, entrepreneurs, and decision-makers, from academia, industry, and government to engage in artificial intelligence, spatiotemporal data computing, and geospatial data science research, motivated by various challenges. In this article, we revisit and discuss the state of GeoAI's open research directions, the recent developments, and an emerging agenda calling for continued cross-disciplinary community engagement.*

## 1 Introduction

We continue observing the field of artificial intelligence (AI) experience significant adoption in new academic programs, new industry offerings, new regulations, and use in everyday broader public spaces. This advancement can be attributed to several factors, including breakthroughs in machine learning (particularly deep learn- ing), the availability of large volumes of data, new hardware architectures and accelerators, and a significant investment in open-source software tools for more accessible data processing, computing, and a higher level of reproducibility. Through many deep learning advances, AI has provided novel solutions to various challenging problems, ranging from computer vision to natural language processing.

The impact of deep learning has reached many application domains. Earth science domains are increasingly transforming due to deep learning methods [46]. Camps-Valls et al. [4] provide a comprehensive review of remote sensing, climate science, and geosciences' open research problems that have been impacted by deep learning advances. Information extraction in remote sensing images has received the most attention, with object detection, classification, and semantic segmentation applications claiming a ”lion’s share” [75, 5, 73]. Generative adversarial networks (GANs) and unsupervised feature representational learning are emerging as powerful tools to augment training data and exploit unlabeled data, respectively [74]. In addition, deep learning methods are equally emerging and showing the transformative capability to detect extreme weather patterns and improve Earth system predictability [1], the adopting of physics-guided learning for parametrization of subgrid processes in climate models [45], as well as bridging the gap between geophysics and geology with GANs [51].

Researchers have also used deep learning techniques from Earth science domains to extract geospatial information from other data sources, including Google Street View [52] or scanned historical maps [13]. In natural language processing (NLP), deep learning models, e.g., those based on recurrent neural networks or Transformers, have improved the accuracy of place name extraction from textual data [36, 59, 24]. Other neural network-based NLP techniques, such as word embeddings, have been employed in studies closer to human geography, e.g., to help quantify changes in stereotypes and attitudes toward women and ethnic minorities over a 100-year study period in the United States [16]. And there are many other examples of research that integrates geography and AI, e.g., for extracting building footprints using convolutional neural networks (CNNs) [67, 71], deep semantic segmentation for automated driving [48], vehicle trajectory prediction [65], indoor navigation [29], gazetteer conflation [47], spatial epidemics [61], and more recently even a demonstration of deep fake geographic maps - where geospatial data is manipulated with AI tools [74]. This continued integration of geography and AI has given rise to the interdisciplinary field of geospatial artificial intelligence (GeoAI) [21].

**Large amounts of geospatial data discoveries with artificial intelligence** and **accelerated hardware/computing platforms** are contributing to the emergence of GeoAI as a field. The massive amount of geospatial data generated by Earth-observing satellites and ground-based sensors offers huge potential for addressing grand societal challenges relating to natural disasters, health, transportation, energy, and food security. Computational methods based on artificial intelligence techniques are more broadly adapted for addressing domain-specific problems. The rise of domain-aware learning is one example that has drawn the in- interest of the GeoAI community. Taking inspiration from the well-known first law of geography [57]: *everything is related to everything else, but near things are more related than distant things*, Deng et al. [11] developed a geographic knowledge-guided neural network, aiming to improve performance in overhead image segmentation tasks. The approach is tailored for when a limited amount of training data is available or when a neural network is applied to test conditions other than which it was trained upon. Continued advances in hardware and computing platforms make it possible to train and deploy complex GeoAI models. Using distributed computing frameworks such as Apache Spark, it is now possible to harness accelerated training and inference workflows. For example, Lunga et al. [34] proposed a remote sensing imagery data flow to improve model generalization across large geographic extents, while an operational large-scale deep learning inference workflow is presented for semantic segmentation [35].

In the above context, a series of GeoAI workshops have been organized at ACM SIGSPATIAL, the premier conference at the intersection of geospatial data analysis and computer science. GeoAI 2017, GeoAI 2018, and GeoAI 2019 took place in Los Angeles, Seattle, and Chicago, respectively [39, 19, 15], while GeoAI 2021 was scheduled to take place in Beijing, China. Due to the COVID-19 pandemic, the workshop was held online (with presentations in a complete virtual format). Since its inception, the GeoAI series of workshops has sought to bring together GIScientists, computer scientists, engineers, entrepreneurs, and decision-makers, from academia, industry, and government to discuss the latest trends, successes, challenges, and opportunities in this new interdisciplinary field. In all previous editions, the GeoAI workshops were among the most popular workshops at the ACM SIGSPATIAL conference, based on conference-reported statistics. The papers accepted and presented at the workshops have covered many GeoAI research topics. In a previous article, we systematically reviewed and summarized the contributions in GeoAI'17, 18, and 19 [18]. In this article, we summarize the research contributions in GeoAI'21, and then describe the research topic evolution at the GeoAI workshops. Finally, we discuss an expanding open agenda in GeoAI research by identifying under-explored directions and calling for broad multi-disciplinary community engagements.

## 2 Research Contributions in GeoAI'21

Due to the hardships and impacts of the COVID-19 pandemic, a 2020 edition of the workshop series was not organized at the ACM SIGSPATIAL 2020 conference, originally planned to take place in Seattle. In 2021, the ACM SIGSPATIAL GeoAI workshop was hosted virtually due to the ongoing constraints posed by the COVID- 19 pandemic. The workshop received 19 submissions, and ten papers were accepted after a rigorous review process. Professor Devis Tuia (Environmental Computational Science and Earth Observation Laboratory, EPFL) gave an academic keynote on "GEOAI in Remote Sensing: Exploiting Natural Language to Understand the Earth", and Dr. Robert Nendorf (Director of Data Science, Arity) and Collin Bennett (Data Engineering, Arity), gave an industry keynote on "Detecting driving behaviors and crashes at 500 trips per second". The accepted papers for GeoAI'21 were mostly distributed across five themes, including methods and techniques (six papers), social media and geo-text analysis (one article), novel and visionary applications (one paper), GeoAI platforms and systems (one paper), and data generation (one paper).

On the theme of methods and techniques, Levering et al. [28] proposed a cross-modal learning strategy to test data and models for the recognition of housing quality in the city of Amsterdam from ground-level and aerial imagery. Gurav et al. [17] presented an approach to conflate geospatial POI data and ground-level imagery while leveraging link prediction on a joint semantic graph. Chen et al. [7] proposed a class-aware unsupervised domain adaptation technique for the semantic segmentation of aerial images. Bowman et al. [3] proposed a method to enable few-shot learning of post-disaster structure damage assessment, leveraging lightweight models and a few labeled samples. Rahman et al. [43] presented a deep learning approach to automatically map road safety barriers from street view imagery, considering road barriers as long objects spanning consecutive street view images in a sequence and adapting a hybrid object-detection and recurrent-network model. Wozniak et al. [60] proposed the hex2vec representation learning method, based on the skip-gram model with negative sampling, aiming to produce context-aware embeddings of H3 hexagons with a basis on OpenStreetMap tags.

On social media and geo-text analysis, Kravi et al. [22] presented a pipeline for geosocial location classification, leveraging machine learning techniques and enabling the association of a set of tweets posted in a small radius around a given location with the corresponding location type.

On the theme of novel applications and visions, Rao et al. [44] proposed a privacy-preserving vehicle trajectory simulation and visualization platform using deep reinforcement learning. The method aims to mitigate data privacy, sparsity, and imbalance sampling issues.

On the theme of GeoAI platforms and systems, Iyer et al. [20] presented early insights into a standardized platform named Trinity that supports complex spatio-temporal problems through multidisciplinary engagements between domain experts, scientists, and engineers. The platform takes a no-code artificial intelligence approach, aiming to enable machine learning researchers and non-technical geospatial domain experts to experiment with domain-specific signals and datasets.

Addressing the issue of data scarcity, as well as the labor-extensive and time-consuming costs of manually annotating text regions in map images, Li et al. [32] proposed style transfer models to convert contemporary map images into historical style while leveraging existing geographic data sources to automatically generate an unlimited amount of annotated historical map images for training text detection models.

## 3 Research Topic Evolution at the GeoAI Workshops

To demonstrate how the research topics in the GeoAI workshops have evolved over the past five years, we created word clouds based on the abstracts in the past three GeoAI workshops (Figure 1). Further, we manually grouped the papers from the past four GeoAI workshops according to their common research themes (Table 1). As can be seen, while the majority of papers focused on geospatial image processing and transportation modeling for the 2017, 2018, and 2019 editions, the 2021 workshop saw a shift toward methods and techniques, as well as the emergence of several new topics that sparked quite an interest (i.e., GeoAI platforms and systems, data generation techniques, data conflation, and location intelligence).

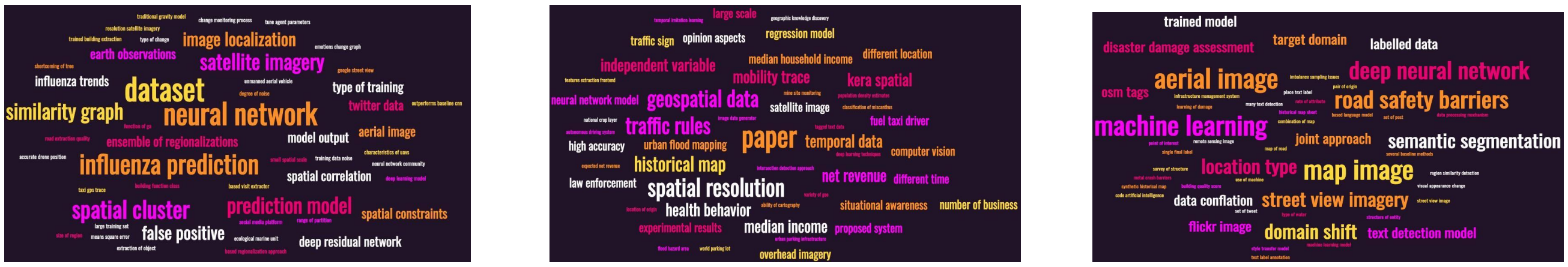

(a) 2018 (b) 2019 (c) 2021

Figure 1: Evolution of GeoAI abstract topics at ACM SIGSPATIAL since 2018.

The popularity of methods and techniques can be attributed to the pervasive outcomes of deep learning-based tools across various domain applications across the ACM SIGSPATIAL community. Other topics addressed in the workshop series include digital humanities, privacy-preserving data mining, cartography, public health, and disaster response. Contrary to 2017, 2018, and 2019, where most studies were based on adapting existing AI methods, papers from 2021 represent an emergence of new methodological research, which indicates tremendous progress in integrating AI tools and acquisition of architectural design knowledge by the ACM SIGSPATIAL community.

# 4 An Expanding Open Agenda in GeoAI Research

The increasing set of different applications, mostly motivated by humanitarian needs, together with the increasingly-prevalent AI tools and their profound impact in society, collectively point to the need for concerted research focusing on new frontiers for GeoAI. With the ever-growing collections of geospatial data, we also anticipate emerging new GeoAI technological advances. These advances should transform GeoAI from a bespoke solution used on certain narrowly-defined domain applications into a commodity technology deployed broadly across GIScience and Earth Science applications. The ACM SIGSPATIAL GeoAI workshop series is one exemplary platform whose goal is to continue fostering a geospatial artificial intelligence community dedicated to resolving problems of greater impact on society.

The workshop envisions a continued engagement, focusing on emerging grand challenges complementary to those being advanced independently by machine learning, geosciences, and remote sensing communities. In the following paragraphs, we ask broad questions motivated by an open agenda and list possible future directions for GeoAI research.

## 4.1 Climate change

Research directions and questions motivated by the climate crisis include:

- Developing GeoAI tools at the intersection of machine learning and climate change crisis.
  - Building applications for agriculture and good security
  - Prototyping tools for carbon capture and sequestration
  - Enhancing methods for climate science and climate modeling
  - Building GeoAI scalable workflows to assist in disaster management and relief
  - Enhancing systems for Earth observation and monitoring
  - GeoAI systems for shading insight in societal adaptation and resilience

Table 1: R&D Proceedings at ACM SIGSPATIAL GeoAI Workshops Series.

| Research Topics | GeoAI Workshop Proceedings 2017 | 2018 | 2019 | 2021 |
|---|---|---|---|---|
| Geospatial image processing | Li, W. et al.[30]<br>Law, S. et al.[27]<br>Collins, C.B. et al. [9]<br>Duan, W. et al. [13] | Xu, Y. et al.[64]<br>Sun, T. et al. [54]<br>Srivastava, S. et al. [52] | Chen et al. [6]<br>Dorji et al. [12]<br>Law et al. [26]<br>Liang et al. [33]<br>Xin et al. (2019) [62] | |
| Transportation modeling and analysis | Kulkarni, V. et al. [25]<br>Murphy, J. et al.[40]<br>Li, Q. et al.[29] | Sun, T. et al. [54]<br>Van Hinsbergh, J. et al [58]<br>Pourebrahim, N. et al [42] | Yin et al. [70]<br>Krumm, J. et al.[23]<br>Xing et al.[63]<br>Yin et al.[69]<br>Mai et al.[37] | |
| Digital humanities | Duan, W. et al. [13] | | Tavakkol et al. [56] | |
| Public health | | Xi, G. et al. [61] | Yang et al. [68] | |
| Disaster response | | | Peng et al. [41] | |
| Social media and geo-text analysis | | Pourebrahim, N. et al[42]<br>Elgarroussi, K., et al.[14] | Yuan et al. [72]<br>Snyder et al. [49] | Kravi, E. et al.[22] |
| Methods and techniques | | Swan, B. et al. [55]<br>Aydin, O. et al. [2] | Soliman et al. [50] | Levering, A. et al. [28]<br>Gurav, R. et al. [17]<br>Ying, C. et al. [7]<br>Bowman, J. et al. [3]<br>Rahman, M. M. et al. [43]<br>Woźniak, S. et al. [60] |
| Novel applications and visions | Majic, I. et al. [38] | Chow, T. E. [8] | Li and Huang [31] | Rao, J. et al. [44] |
| GeoAI platforms and systems | | | | Iyer, C. V. K. et al. [20] |
| Data generation | | | | Chen, Y. et al. [32] |

## 4.2 Ethics and trust in GeoAI

Research directions and questions motivated by the need for reliable and explainable models include:

- Building robust and reliable GeoAI models. How can we ensure that future GeoAI systems are robust and reliable, and how do we evaluate them at scale?
- Designing and prototyping explainable GeoAI models. Most AI learning systems remain a black box, especially for domain scientists looking for AI-enabled solutions. While these systems have demonstrated good performance in many tasks, such as object detection and image classification, it is equally important to understand their learning and engage with decision-makers when applied to various geospatial data analysis problems.

## 4.3 Methods for multimodal learning and generalization

Research directions and questions motivated by the uniqueness of geospatial generalization problems and phenomena include:

- Building spatiotemporally explicit models. To better understand the complex geospatial contexts and geographical processes on the ground, it is crucial to employ spatiotemporally explicit models and evaluate the results by integrating human intelligence and machine intelligence evaluations [66].
- Enhancing model generalizability in the context of geography and sensors. Labeled datasets and imagery are always collected from certain regions with certain sensors. How can we ensure that the GeoAI models trained using one specific data from one geographic area or one type of sensor can be generalized to other new geospatial data?
- Accommodating uncertainty in geospatial problems and datasets. Deep learning methods have traditionally not been designed with data uncertainty in mind, although uncertainty is a fundamental concept in geography. Can geography contribute to deep learning more broadly by developing methods to imbue the models with uncertainty analysis?
- Fusing multi-source geospatial data for knowledge discovery. Fusing diverse geospatial datasets at different Spatio-temporal resolutions and through feature engineering and deep learning can enable novel geographic knowledge. How can machine learning help automate, streamline, or assist geospatial data integration?

## 4.4 Democratization of GeoAI data, tools and platforms

Research directions and questions motivated by the wealth of geospatial data include:

- Developing new and open geospatial data infrastructures. ImageNet [10] played a key role in revolutionizing the field of computer vision. Future GeoAI applications could benefit from similar platforms by investigating open and indexable rich geospatial data archives. A great example is a BigEarthNet platform [53], which has been significantly larger than other archives in remote sensing and has been used as a diverse training source in deep learning.
- Building domain datasets via novel techniques. As most GeoAI models involve supervised learning, the availability of large benchmark datasets becomes essential to promote research across geospatial communities and validate the research results' generalizability. Can we leverage novel techniques to facilitate the development of domain datasets? For example, is there a role for GANs to play in augment- ing GeoAI training data?
- Reducing the need for labeled data through self-supervised learning. Self-supervised learning methods have demonstrated their capability in other domains with their potentially unlimited ability to uncover patterns in unlabeled data. Could they offer better scalability and reduce the need for labeled training data in GeoAI applications?

The aforementioned research directions and questions are not exhaustive, and many other important directions are to be explored. We look forward to seeing exciting new research to be shared and published in the ACM SIGSPATIAL GeoAI workshop series, as well as in other related venues, in the coming years.

## 5 Workshop Organization

ACM SIGSPATIAL GeoAI'21 was led by the organizing committee (Dalton Lunga, Lexie Yang, Song Gao, Bruno Martins, Yingjie Hu, Xueqing Deng, Shawn Newsam).

- **Program Committee:**
  - **Pete Atkinson**, Atkinson, Lancaster University, UK
  - **Orhun Aydin**, Esri Inc., USA
  - **Booma S. Balasubra- mani**, Microsoft, USA
  - **Dengfeng Chai**, Zhejiang University, China
  - **Yao-Yi Chiang**, University of Southern California, USA
  - **Xiao Huang**, University of Arkansas
  - **Zhe Jiang**, University of Alabama
  - **Kuldeep Kurte**, Oak Ridge National Laboratory, USA
  - **Wenwen Li**, Arizona State University, USA
  - **Xiaojiang Li**, Temple University, USA
  - **Yanhua Li**, Worcester Polytechnic Institute, USA
  - **Gengchen Mai**, Stanford University, USA
  - **Claudio Persello**, University of Twente, Netherlands
  - **Devis Tuia**, EPFL, Switzerland
  - **Martin Werner**, Technical University of Munich, Germany
  - **Yiqun Xie**, University of Mary- land, College Park, USA
  - **Fan Zhang**, MIT Senseable City Lab, USA
  - **Di Zhu**, Peking University, China
  - **Lei Zou**, Texas A&M Univer- sity, USA